\documentclass[sigconf]{acmart}
\usepackage{graphicx}
\usepackage{mathrsfs}
\usepackage{amsfonts}
\usepackage{booktabs}
\usepackage{bm}
\usepackage{epsfig}
\usepackage{graphicx}
\usepackage{amsmath}
\usepackage{amsthm}

\usepackage{amssymb} % define this before the line numbering.
\usepackage{multirow}
\usepackage{bbm}
\usepackage{bbding}
\usepackage{graphics}
\usepackage{caption, subcaption}
\usepackage{float}
\usepackage{cuted}
\usepackage{arydshln}
\usepackage{slashed}
\usepackage{soul}
\usepackage{balance}
\AtBeginDocument{%
	\providecommand\BibTeX{{%
			\normalfont B\kern-0.5em{\scshape i\kern-0.25em b}\kern-0.8em\TeX}}}

\copyrightyear{2022}
\acmYear{2022}
\setcopyright{acmcopyright}\acmConference[MM '22]{Proceedings of the 30th ACM
	International Conference on Multimedia}{October 10--14, 2022}{Lisboa, Portugal}
\acmBooktitle{Proceedings of the 30th ACM International Conference on Multimedia
	(MM '22), October 10--14, 2022, Lisboa, Portugal}
\acmPrice{15.00}
\acmDOI{10.1145/3503161.3547794}
\acmISBN{978-1-4503-9203-7/22/10}

\begin{document}
	
	\title{Set-Based Face Recognition Beyond Disentanglement: \\ Burstiness Suppression With Variance Vocabulary}
	%   Burstiness suppression with variance vocabulary
	
%\author{\mbox{Jiong Wang \hspace{10pt} Zhou Zhao* \hspace{10pt} Fei Wu \hspace{10pt}}}
%\author{\mbox{Jiong Wang \hspace{10pt} Zhou Zhao* \hspace{10pt} Fei Wu \hspace{10pt}}}
%\affiliation{College of Computer Science and Technology, Zhejiang University, China}
%\authornote{Corresponding authors}
%\email{{liubinggunzu, zhaozhou, wufei}@zju.edu.cn}

\author{Jiong Wang}
%\authornote{Dr.~Trovato insisted his name be first.}
\affiliation{%
	\institution{College of Computer Science and Technology, Zhejiang University, China}
	%\institution{Shenzhen University}
	%\streetaddress{P.O. Box 1212}
	% \city{Shenzhen}
	%  \state{China}
}
\email{liubinggunzu@zju.edu.cn}

\author{Zhou Zhao}
\authornote{Corresponding author}
\affiliation{%
	\institution{College of Computer Science and Technology, Zhejiang University, China \\
	Alibaba-Zhejiang University Joint Research Institute of Frontier Technologies}
}
\email{zhaozhou@zju.edu.cn}

\author{Fei Wu}
\affiliation{%
	\institution{College of Computer Science and Technology, Zhejiang University, China}
}
\email{wufei@zju.edu.cn}

\renewcommand{\shortauthors}{Jiong Wang, Zhou Zhao, Fei Wu}
%\renewcommand{\shortauthors}{Jiong Wang, Zhou Zhao, & Fei Wu}
%% If needed use a foot or author note to identify equal contribution
%\renewcommand{\shortauthors}{Jiong Wang, Zhou Zhao, Fei Wu et al.}	
	
	%% The abstract is a short summary of the work to be presented in the article.
	\begin{abstract}
		Set-based face recognition (SFR) aims to recognize the face sets in the unconstrained scenario, where the appearance of same identity may change dramatically with extreme variances (e.g., illumination, pose, expression). We argue that the two crucial issues in SFR, the face quality and burstiness, are both identity-irrelevant and variance-relevant. 
		The quality and burstiness assessment are interfered with by the entanglement of identity, and the face recognition is interfered with by the entanglement of variance. Thus we propose to separate the identity features with the variance features in a light-weighted set-based disentanglement framework. 
		%Facilitated by disentanglement, the identity features are robust to the variances.
		
		Beyond disentanglement, the variance features are fully utilized to indicate face quality and burstiness in a set, rather than being discarded after training. 
		%We adopt the variance features to indicate face quality attention, and is better than entangled features. 
		%The disentangled variance features are 
		To suppress face burstiness in the sets, we propose a vocabulary-based burst suppression (VBS) method which quantizes faces with a reference vocabulary. With inter-word and intra-word normalization operations on the assignment scores, the face burtisness degrees are appropriately  estimated. 
		%The proposed VB aggregation overcomes the drawbacks of existing burtsiness suppression methods and considerably improve the recognition performance. 
		The extensive illustrations and experiments demonstrate the effect of the  disentanglement framework with VBS, which gets new state-of-the-art on the SFR benchmarks. The code will be released at \url{https://github.com/Liubinggunzu/set_burstiness}.
		
	\end{abstract}
	
	
	
	%\ccsdesc[300]{Computing methodologies~Object identification}
	
	 %Burstiness Analysis; Convolutional Neural Network}
	
	\maketitle
	
	\begin{figure}[t]
		\begin{center}
			\includegraphics[scale=0.43]{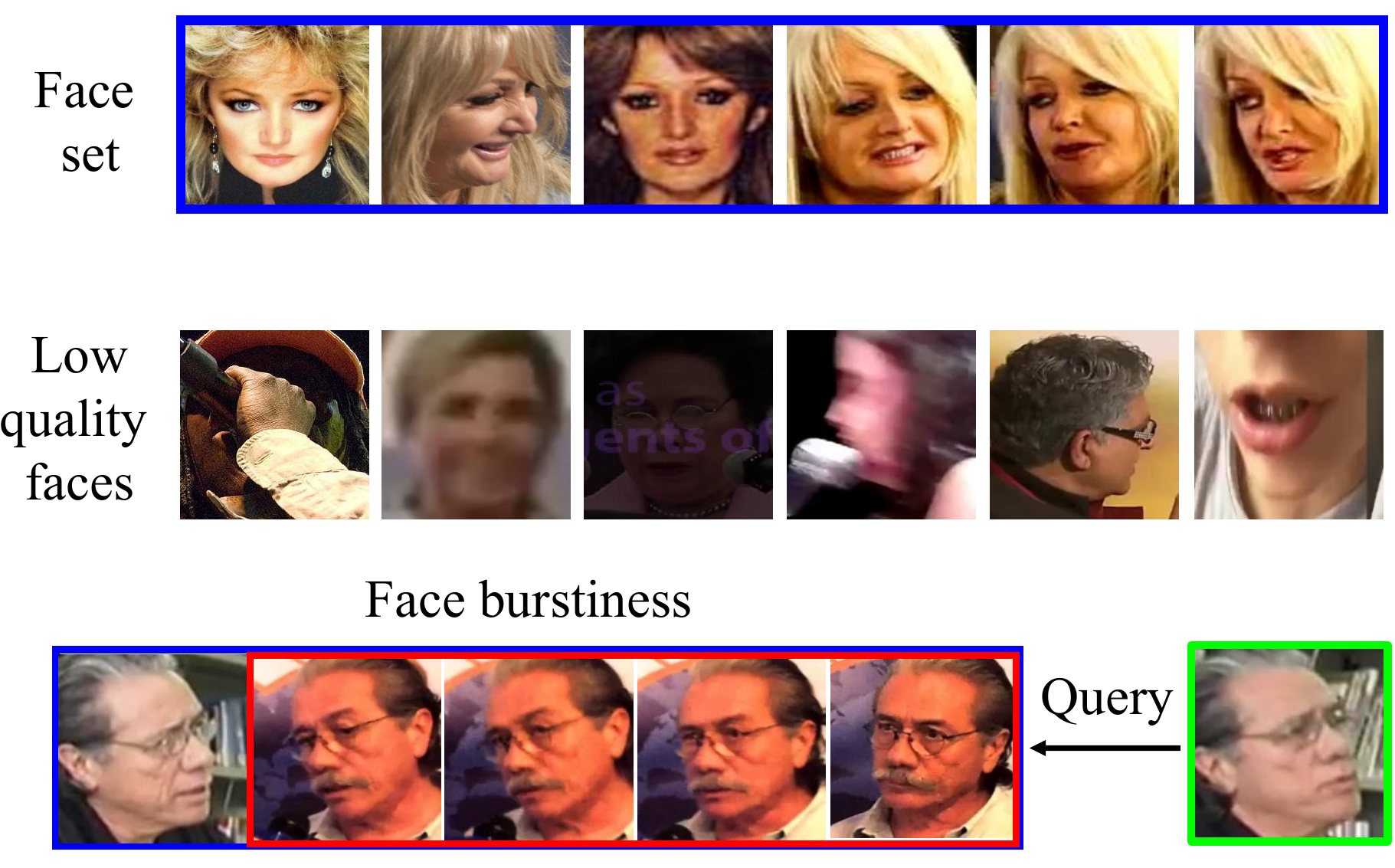}\vspace{-0.2cm}
		\end{center}
		% \vspace{6pt}
		\caption{(First row) The face set consists of faces of the same identity with sundry variations. (Second row) Exemplary low-quality faces in the unconstrained scenario. (Third row) Illustration of the burstiness problem, where the set representation is dominated by frequent face features and undemocratic to all the variations in the set.   
		}
		\label{fig:motivation}
		\vspace{-0.2cm}
	\end{figure}
	
	\section{Introduction}
	%heuristic solution. Even the get performance drop on the varying condition changes, Even when the face set may be easily by particular pose when they are dominated. Our goal is not biased to particular . 
	
	Set-based face recognition (SFR) requires discriminative and compact representation to characterize the face set (First row of Figure~\ref{fig:motivation}), which may contain faces of the same identity with varying cardinality (from one to hundreds or more) and from multiple sources (\emph{e.g.}, static images, video frames, or a mixture of both).
	SFR has attracted consistent interest in the computer vision community as a growing number of both face images and videos are continually uploaded to the Internet and being captured by the terminal devices. Storing and searching all the face features are unpractical, thus it is necessary to design compact and representative set representations. 
	%makes SFR a necessary way to get compact and representative set representations. 
	%(it is n to use compact represent)
	
	Because of the unconstrained capture environments in terms of varying poses, illuminations, motion blurs and occlusions, it is challenging for a SFR system to get reasonable set representations under various face quality conditions. Some examples of the low-quality faces are illustrated in the second row of Figure~\ref{fig:motivation}. 
	% Compared to single face recognition, a compact set-descriptor to speed of face recognition. SFR in the unconstrained scenario is a challenging
	%The application of Set-based face recognition is . (You can refer to the app of single image)
	Even though the face recognition performance is dramatically improved in the past decade, %facilitated by the success of deep learning. 
	recognizing faces with variance conditions in the unconstrained scenario is still challenging \cite{sengupta2016frontal, moschoglou2017agedb, zheng2017cross, zheng2018cross}. 
	%Having the prior that faces with extreme conditions are ill-learned, 
	Under such a circumstance, face quality attention is widely studied in existing SFR works \cite{yang2017neural, liu2017quality, xie2018multicolumn, xie2020inducing} to suppress the low-quality faces while strengthening the high qualities.  
	
	However, the robustness of face features is still limited by the entanglement of identity and variances. In this paper, we propose to separate the identity features and variance features with a light-weighted set-based disentangled representation learning (DRL) framework. 
	Similar to existing disentanglement works \cite{zhang2019gait, peng2017reconstruction}, the disentangled identity features are discriminative and robust to variances. Beyond these works,
	the set representation is involved in the proposed set-based DRL process and is expected to be discriminative and representative.
	%the proposed DRL framework is set-based with the set representations involved in the DRL process. 
	In addition, the variance features are fully utilized to indicate face quality and burstiness, rather than being discarded. %\textbf{In addition, the proposed DRL is set-based.}

	Furthermore, we argue that existing works ignore a crucial factor in face sets, the burstiness phenomenon \cite{madsen2005modeling, jegou2009burstiness}, where frequent faces with particular attributes dominate the set. In an ideal case, the set representation should be democratic \cite{jegou2014triangulation} to faces of all the variations exhibited in a set for the generalization ability on the unconstrained scenario.  While set representation with sum-aggregation naturally focuses on the frequent faces and loses the generalization ability to un-frequent faces in the set. As shown in the third row of Figure~\ref{fig:motivation}, a database face set is composed of bursty faces (red border) and a non-bursty face. A query face (set) with similar attributes to the non-bursty face usually gets lower similarity to the sum-aggregated database representation. 
	
	To suppress the burstiness phenomenon, Shi \emph{et al.} \cite{shi2015early} propose to detect feature groups with clustering algorithms, but it is not applicable to small sets. J{\'e}gou \emph{et al.} \cite{jegou2014triangulation, murray2016interferences} and Murray \emph{et al.} \cite{murray2014generalized, murray2016interferences} separately propose the democratic aggregation (DA) and the generalized max-pooling (GMP) with a similar objective to equalize the similarity between each element feature and the set representation. These two methods based on the analysis of self-similarity matrix, however, are interfered with by low-quality faces, and we found they hardly take effect on the unconstrained benchmarks.
	
	To overcome the drawbacks of existing burst suppression works, we propose a vocabulary-based burst suppression (VBS) method, which is based on the analysis of reference-similarity matrix from the face set to a variance vocabulary. 
	The face variance features are quantized by the vocabulary and the assignment scores are expected to capture the burstiness information. The inter-word and intra-word normalization operations on the assignment scores are devised to constrain the burstiness effect intra the words, where bursty words are relatively suppressed. %Compared to existing burstiness suppression methods, 
	We give extensive experiments to demonstrate the VBS method is applicable to various scenarios with prominent effectiveness and efficiency.

	The contributions are summarized as follows. (1) We propose a light-weighted set-based disentanglement framework for SFR task, and emphasize the usage of variance features to solve the face quality and face burstiness problem. 	(2) We propose a vocabulary-based burst suppression (VBS) method to assess the burst degree of faces in the set. The VBS is more effective, efficient than existing burstiness suppression works, and they are also complementary on the YTF dataset. (3) We give detailed illustrations and qualitative results to dissect the proposed VBS. The extensive experiments demonstrate the effectiveness of disentanglement with vocabulary-based aggregation, which gets new state-of-the-art results on SFR benchmarks. 
	
	\section{Related Works} \label{sec:related}
	
	\subsection{Set-based Face Recognition}
	Set-based face recognition (SFR) has been widely studied and the earlier works represent the face set as manifold \cite{lee2003video, arandjelovic2006information, huang2015projection, wang2015discriminant}, convex hull \cite{cevikalp2010face, cevikalp2019discriminatively} or set covariance matrix \cite{wang2017discriminative}, 
	and measure the set distance in corresponding space. 
	%Deep face features have shown the discrimination ability and compactness for representing faces. 
	The deep face recognition models \cite{schroff2015facenet, taigman2014deepface, cao2018vggface2, deng2019arcface} recently demonstrate prominent performance on the SFR benchmarks with a simple sum-aggregation strategy.
	Face quality attention is an effective strategy to improve the sum-aggregation and is widely studied in existing SFR works \cite{yang2017neural, liu2017quality, xie2018multicolumn, gong2019video, xie2020inducing, zhang2020discriminability} to weaken the contribution of low-quality faces while strengthening the high qualities. %VLAD \cite{jegou2010aggregating, zhong2018ghostvlad} is also adopted to aggregate the face features to a compact representation.  
	
	However, the feature robustness is still limited and the burstiness problem is ignored by these works. We propose to separate identity features with variance features in a light-weighted set-based disentanglement framework. The disentangled identity features are more discriminative and the variance features are more representative to face quality and burstiness than the entangled features. 
	%We also use VA to indicate face burstiness with VB aggregation.
	%We also propose a VB aggregation to suppress the burstiness phenomenon, ... follow the writing in introduction.
	
	\begin{figure*}[t]
		\begin{center}
			\includegraphics[scale=0.45]{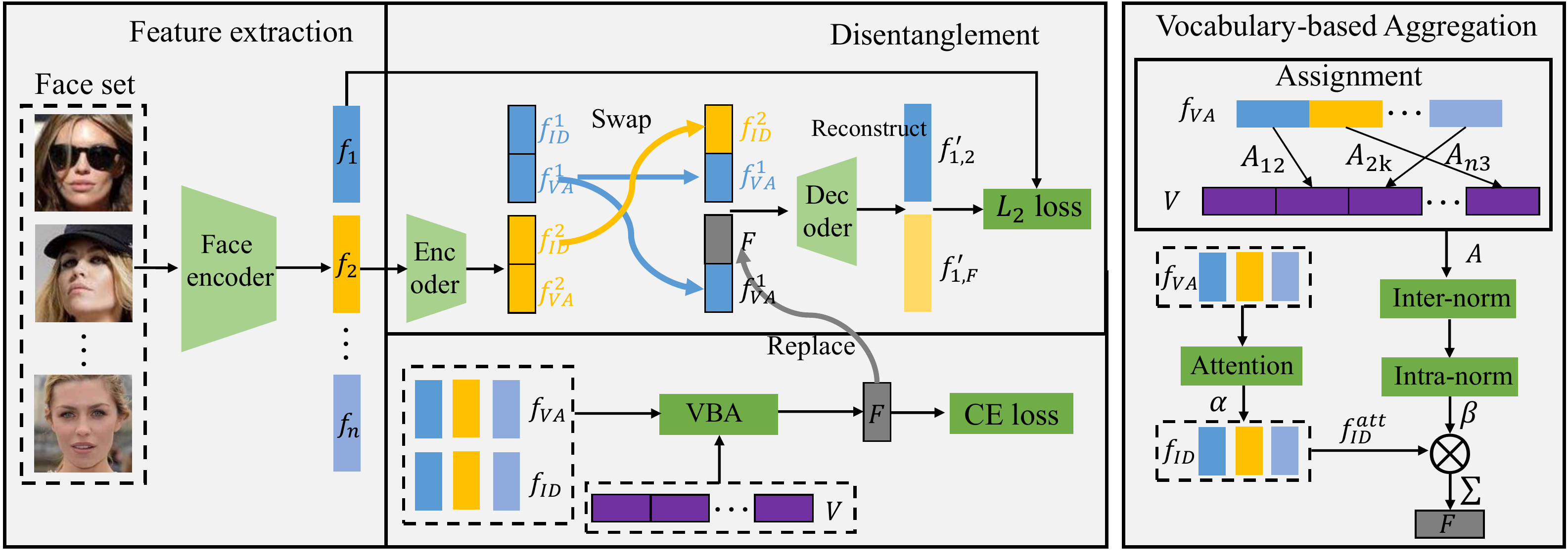}\vspace{-0.2cm}
		\end{center}
		% \vspace{6pt}
		\caption{Illustration of the light-weighted set-based disentanglement framework with vocabulary-based aggregation. The set-based disentanglement additionally replaces the original identity features with set representation for reconstruction. The vocabulary-based aggregation (VBA) assesses the face quality and burstiness with the disentangled variance features.
		}
		\label{fig:pipeline}
		\vspace{-0.2cm}
	\end{figure*}

	\subsection{Visual Burstiness Analysis}
	Inspired the burstiness analysis in text recognition \cite{church1995poisson, madsen2005modeling, he2007using}, the burstiness phenomenon was brought to attention in former image retrieval works \cite{torii2013visual, jegou2009burstiness, shi2015early}, where the repetitive structures in natural images lead to burstiness and corrupt the visual similarity measurement. %Shi \emph{et al.} \cite{shi2015early} propose to detect bursty groups with clustering algorithms, the mean-shift \cite{cheng1995mean, vedaldi2008quick} or $k$-means, and equalize all the groups by normalizing the group features. 

	Derived from similar inspiration to equalize the similarity between each
	element feature and set representation,  J{\'e}gou \emph{et al.} \cite{jegou2014triangulation, murray2016interferences} and Murray \emph{et al.} \cite{murray2014generalized, murray2016interferences} separately propose the democratic aggregation (DA) and generalized max-pooling (GMP). DA \cite{jegou2014triangulation} adopts a modified Sinkhorn algorithm to iteratively calculate the democratic weights, while GMP \cite{murray2014generalized} calculates the burst weights by solving the linear-regression problem. %Comparatively, GMP performs better in the comparison of \cite{murray2016interferences}.

	However, %the method in \cite{shi2015early}  is not applicable to small sets and suffers from high computation overhead when applied on large sets. 
	the GMP and DA also suffer from the complexity problem on large sets because the complexity of gram matrix is $\mathcal{O}(n^2d)$. %and the complexity of following iteratively matrix multiplication or matrix inverse is higher. 
	Moreover, the GMP and DA, as two self-similarity-based methods, are interfered with by the low-quality faces in the unconstrained scenario. The low-quality face features with lower discriminability are usually highlighted because of their lower self-similarities in the set.  We therefore propose the VBS, a reference-similarity-based method, with analysis on the assignment matrix of feature set to the variance vocabulary.
	VBS has the complexity of $\mathcal{O}(nkd)$ and is applicable on the unconstrained scenario with prominent effects.
	
	%But we found GMP and DA are not applicable on SFR  because they are based on the self-similarity and thus are interfered by the low-quality faces in the unconstrained scenario. The proposed vocabulary-based aggregation is advanced in three aspects: Firstly, it is based on the referenced similarity and is robust to low-quality faces.  Secondly, it is end-to-end optimized with pronominent performance. Thirdly, it is applicable to sets of any size and don't suffer from much computation problem.
	
	%The proposed Dem loss derive similar objective with GMP and DA, and here it is a transformed constraint in the training stage.
	
	%The proposed VB is inspired by to quantize, but different from clustering with mean-shift, we propose to quantize with disentangled VA vocabulary, which is computation free and applicable to sets of any size. Compared to GMP and DA, it possesses efficiency, effectiveness, and scalability on the online-updating of a set.   
	
	%We propose a vocabulary by the disentangled variance features. The is advanced in three aspects:  First, disentangled variance is complementary to GMP and DA method. Second, vocabulary is advanced in efficiency and scalability: Vocabulary by normalization and don't need to calculate the gram matrix. Second, Vo is applied to the dynamic updating the database faces, while GMP and DA requires to re-calculate the gram matrix when introducing new faces to database.
	
	\subsection{Disentangled Representation Learning}
	Disentangled representation learning (DRL) has been previously studied on the bilinear models \cite{tenenbaum2000separating}, restricted Boltzmann machines \cite{desjardins2012disentangling, reed2014learning} and variational autoencoders \cite{chen2018isolating, higgins2016beta, kim2018disentangling, kingma2013auto}.
	The notion of DRL is recently applied on various multi-media recognition tasks, such as object detection \cite{lin2021domain, chen2021disentangle, wu2021vector}, gait recognition \cite{zhang2019gait, li2020gait}, %facial expression analysis \cite{koujan2020real, zhang2021learning} 
	and person re-identification \cite{eom2019learning}  with similar goals to separate the distraction of variances.
	%and visual generation \cite{nguyen2019hologan, deng2020disentangled, }. 
	Disentangled face recognition is also studied for cross-pose \cite{tran2017disentangled, peng2017reconstruction} and cross-age \cite{zhao2019look, wang2019decorrelated} face recognition tasks.  
	
	%Different from these works, we adopt a light-weighted set-based DRL framework where the identity features and variance features in a set are swapped each other and reconstruct the original entangled features. 
	
	Inspired by these works, %where the identity features and variance features in a set are swapped with each other and reconstruct the original images.  
	we propose a light-weighted set-based DRL framework and the differences are in three aspects: First, the proposed framework is \textbf{light-weighted} and only the original entangled features are reconstructed by a fully-connected layer. Second, the final set representation, which are expected to be discriminative and democratic, is also concatenated with the variance features to reconstruct the entangled features (\textbf{Set-based}).
	Third, the disentangled variance features are fully utilized to indicate face quality and burstiness in the set.  
	%Compared to these works, the proposed disentanglement framework is more advanced in two aspects: We consider propose a light-weight frame work to apply Disentanglement in a face set, and propose a margin loss to guarantee them with desirable characters. Secondly, We make full use of the variance features, rather than discarding them in existing disentangled works. We also use it to indicate face quality an face burstiness in set.

	\section{The Proposed Method} \label{sec:method}
	
	In this section, we first introduce the conventional set-based face recognition (SFR) pipeline (Section \ref{sec:preliminary}) and the concept of burstiness and feature entanglement  (Section \ref{sec:burstiness}).  We then detail the set-based disentanglement process (Section \ref{sec:disentangle}) and the proposed vocabulary-based aggregation method  (Section \ref{sec:vagg}). % We finally describe the IJBC-BS protocol (Section \ref{sec:ijbcbs}). 
		
	\subsection{Preliminary} \label{sec:preliminary}
	
	The conventional SFR pipeline adopts attention-aware aggregation, where the attention block is essential in existing SFR works \cite{yang2017neural, gong2019video, xie2020inducing}. 
	Supposing a face set $\mathcal{X} = \{\rm{x}_1, \rm{x}_2, ..., \rm{x}_{n}\}$, composed of $n$ faces with same identity, is passed to the face encoder and the feature set $f = \{f_1, f_2, ..., f_n\}$ is obtained with shape of $n \times d$.
	%In the aggregation module 
	The vanilla attention block \cite{yang2017neural,xie2018multicolumn} consists of a $d$ dimensional query vector $q$ to evaluate the face features and a following sigmoid function to scale the numerical values. %and get the attention scores. 
	Finally, sum aggregation with attention weights gets the $d$ dimensional set representation $F$:
	\begin{equation}
	F = \sum\nolimits_{i=1}^{n} \beta_{i} f_i, \  \  \alpha_{i} = \bm{\sigma} (q f_i^T).
	\end{equation}
	$F$ is optimized for identity classification in the training stage, for face verification and identification in the evaluation stage.
	
	%A vanilla pipeline to get the set representation is illustrated in Figure~\ref{fig:pipeline}, where the faces in a set are passed to the face encoder and the resulted face features are aggregated in the aggregation module to the set representation.  The aggregation module consists of an attention block to weigh each feature and the sum-aggregation to get the set representation. 
	
	%In each training epoch, all the identities in the training set are sampled and 

	%For SFR task, the pre-trained face encoder is assumed to be available and usually the pre-trained dataset is used for fine-tuning \cite{xie2018multicolumn,zhong2018ghostvlad}. So we define the ``fine-tuning'' process as training. 
	
	\subsection{Burstiness and Feature Entanglement} \label{sec:burstiness}
	%discuss quality above.
	
	Another crucial factor apart from face quality in SFR is the face burstiness, a phenomenon that frequent faces with particular attributes dominate the face set.  The sum-aggregated set representations are naturally dominated by the frequent face features and lose generalization ability to the un-frequent faces. Existing Burstiness suppression methods such as DA \cite{jegou2014triangulation} and GMP \cite{murray2014generalized} are based on the analysis of the gram (self-similarity) matrix. 
	
	Note that faces in a set share same identity, face quality and face burstiness in sets are thus identity-unrelated and variance-related  while their assessments suffer from the entanglement of identity features.
	Supposing the original feature entangled with the identity feature and variance feature \cite{zhang2019gait, li2020gait} is formulated as: %and the entanglement are concatenation:
	\begin{equation}
	f = [f_{ID}^{}, f_{VA}^{}].
	\end{equation}
	The face quality and face burstiness are separately indicated by the quality attention score $\alpha$ and the gram matrix $G$: 
	\begin{equation}
	\alpha = \bm{\sigma} (q_{ID}^{} f_{ID}^T + \mathbf{q_{VA}^{} f_{VA}^T}), \ \ q = [q_{ID}^{}, q_{VA}^{}],
	\end{equation}
	\begin{equation}
	G = ff^T = f_{ID}^{} f_{ID}^T +  \mathbf{f_{VA}^{} f_{VA}^T}.
	\end{equation}
	
	It can be seen that the entanglement of identity feature interferes the estimation of these two crucial factors. %, and lead to non-robust results.   
	Therefore we propose to separate the identity feature and variance feature from original entangled feature and utilize the identity feature for the aggregation of set representation, variance feature for quality and burstiness assessment.
	
	\begin{figure*}[t]
		
		\centerline{
			\hspace{0.5cm}
			\subcaptionbox{\small Feature sets}{
				\includegraphics[scale = 0.41]{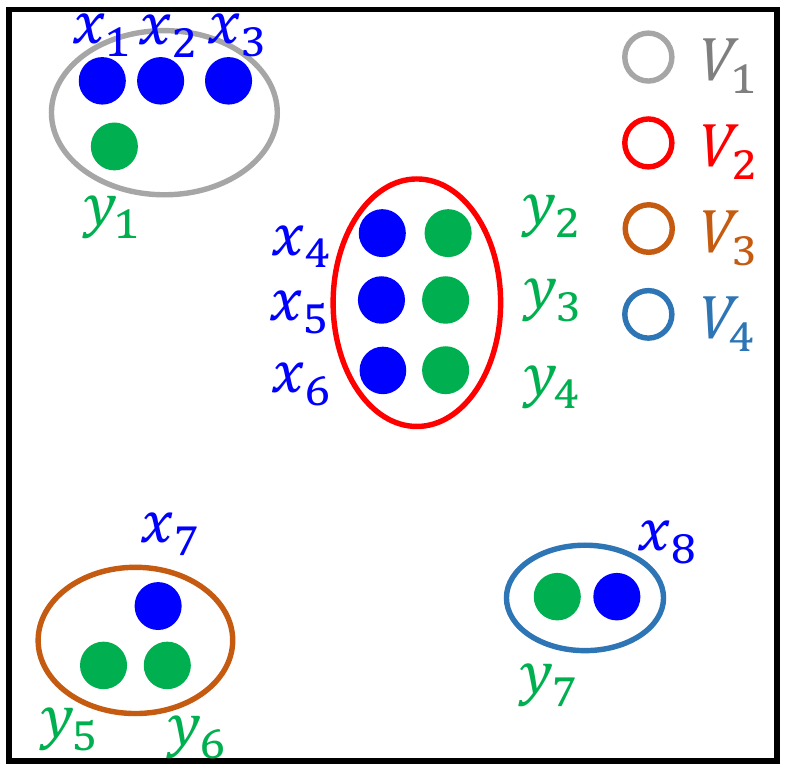} \hspace{0.2cm}
			} \hspace{-0.1cm}
			\subcaptionbox{\small Cross-similarity}{
				\includegraphics[scale = 0.3]{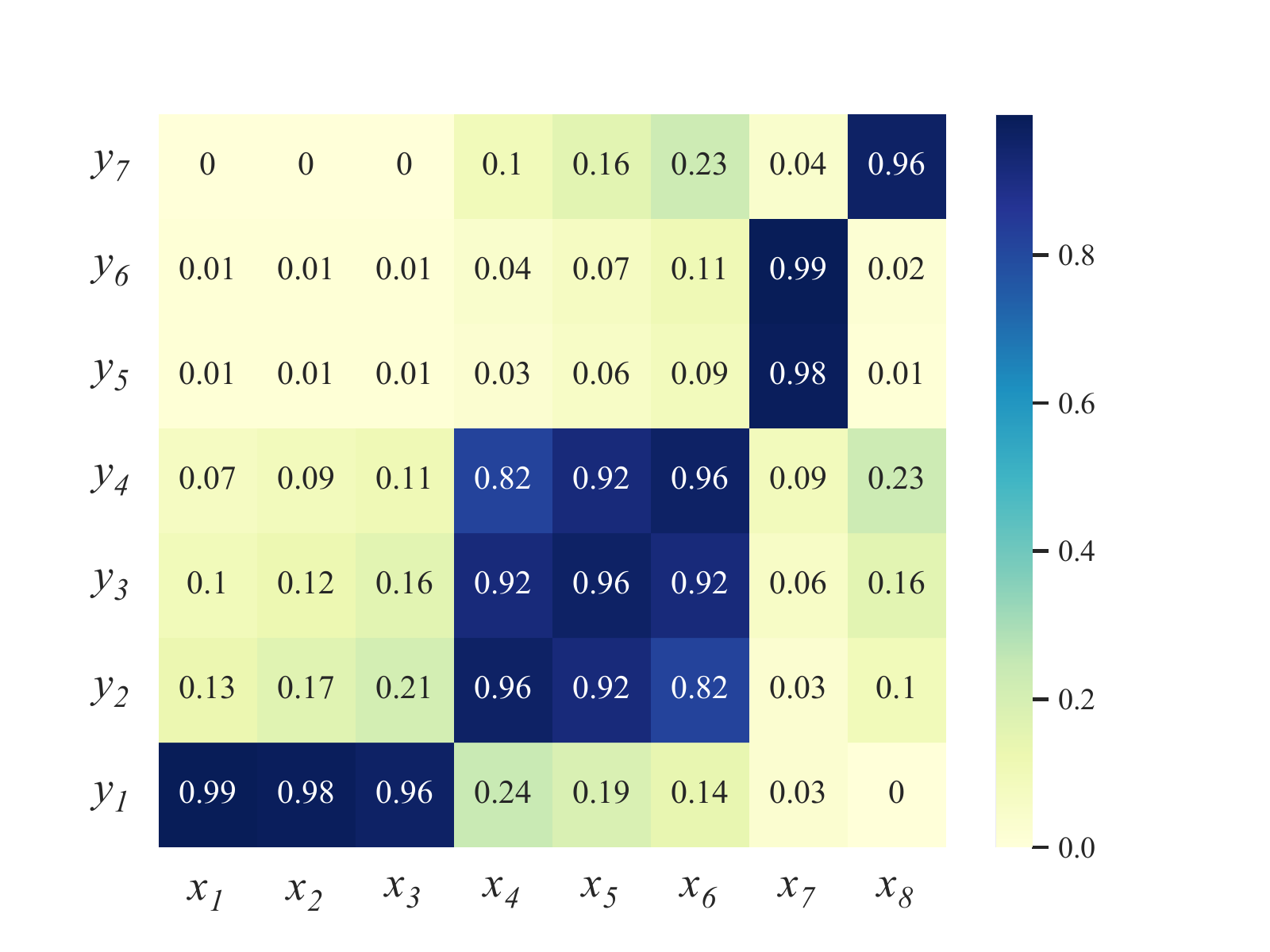} \vspace{-0.1cm}
			} \hspace{-0.3cm}
			\subcaptionbox{\small VBS weights}{
				\includegraphics[scale = 0.45]{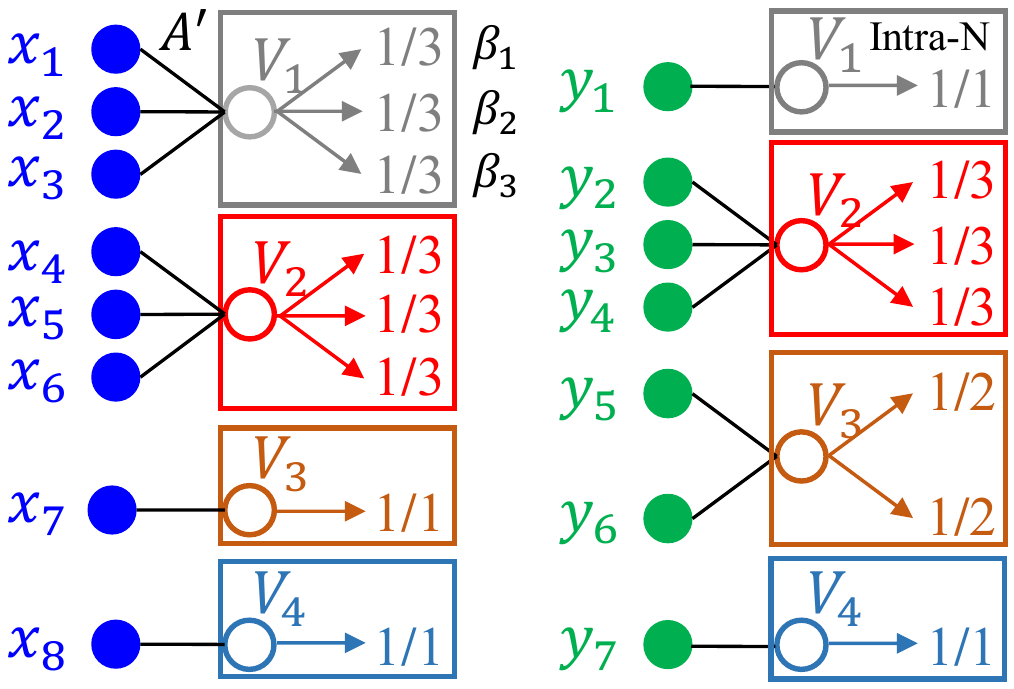} \hspace{0.2cm}
			} \hspace{-0.2cm}
			\subcaptionbox{\small Cross-similarity (VBS)}{
				\includegraphics[scale = 0.3]{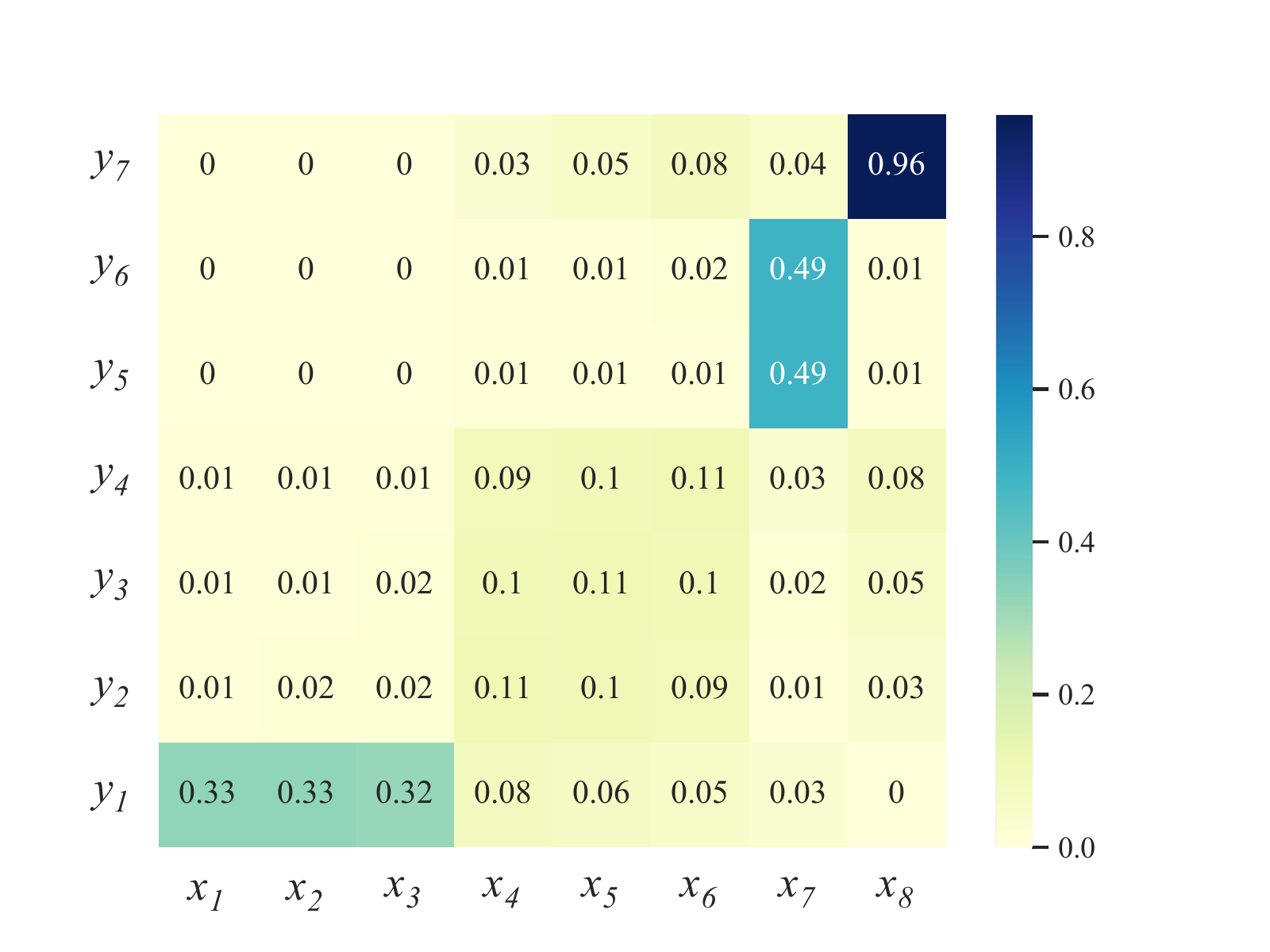} \vspace{-0.1cm}
			} %\vspace{-0.3cm}
			%\subcaptionbox{\small Assignment}{
			%\includegraphics[scale = 0.25]{assign.JPG} \vspace{-0.1cm}
			%} %\vspace{-0.5cm}
		}  %\vspace{-0.2cm}
		\caption{Toy example of two feature sets on the 2D space quantized by a vocabulary. The cross-similarities are dominated by the bursty words and the proposed vocabulary-based burst suppression (VBS) equalizes the cross-similarities of all the visual words with  intra-normalized weights. The similarity in 2D space is transformed from the distance $sim_{i, j} = \exp(-\left \| x_i - y_j \right \|_2)$.
		}
		\label{fig:explain}
		\vspace{-0.2cm}
	\end{figure*}
	\subsection{Set-Based Disentangled Representation Learning} \label{sec:disentangle}
	%To separate the ID features and VA features from original disentangled features.
	
	Inspired by former image-based disentangled representation learning (DRL) works \cite{zhang2019gait, eom2019learning}, which involve swapping the identity and variance features to reconstruct the original images. 
	We propose a light-weighted set-based disentanglement framework with only the original entangled features reconstructed, and the identity features are additionally replaced  by the set representation for reconstruction.
	%additionally replacing identity features with the set representation for reconstruction.
	
	%The proposed set-based DRL framework is light-weighted and only the original entangled features are reconstructed. 
	%We additionally replace the identity features with the final set representation for reconstruction, for the reason the set representation is expected to be discriminative and democratic to all the faces in the set.
	
	%These DRL notion helps the separation of identity and variance features. 
	
	As shown in Figure~\ref{fig:pipeline}, the face features extracted from the face encoder are projected in the disentanglement encoder with two fully-connected (FC) layers to identity features and variance features, which are separately expected to capture the identity and variance information.  
	\begin{equation}\label{key}
	f_{ID} = Enc_{I}(f), \ \ f_{VA} = Enc_{V}(f) .
	\end{equation}
	In each face set, the identity features are swapped each other and concatenated with original variance features to reconstruct original features in the decoder with a FC layer. The interpretation is that the %faces in a set share same identity and 
	identity features in a set are expected to be invariant to variance, and thus the exchanged features can still reconstruct the original features. 
	Following \cite{zhang2019gait, li2020gait}, we adopt $L_2$ loss to evaluate these image-based reconstruction results.  
	\begin{equation} \label{eq:imgd}
	\begin{split}
	L_{recons-img}(f_i, f_j) = \left \| f_i - f_{i,j}^{\prime} \right \|_2, \ \ f_{i,j}^{\prime} = Dec([f^{j}_{ID}, f^{i}_{VA}])
	\end{split}
	\end{equation}
	
	In addition to swapping the identity features, we also replace the original identity features with set representation for reconstruction. The motivation is that the final set representation is expected to be discriminative and democratic for all the faces in the set. 
	\begin{equation} \label{eq:setd}
	\begin{split}
	L_{recons-set}(f_i, F) = \left \| f_i - f_{i,F}^{\prime} \right \|_2, \ \ f_{i,F}^{\prime} = Dec([F, f^{i}_{VA}]), 
	\end{split}
	\end{equation}
	where $F$ is the aggregated set representation as formulated in Equation~\ref{eq:f}.
	
	\noindent\textbf{Discussion.}
	Different from previous works \cite{eom2019learning, zhang2019gait}, the proposed DRL framework is light-weighted and the disentanglement decoder reconstructs original features rather than face images. 
	The reason is that the SFR model converges fast in the training process because the pre-trained face encoder is usually frozen and only remaining parameters are optimized. When a face decoder is adopted to reconstruct the original faces, numerous parameters are brought and we found it hardly converges.  
	%The proposed light-weighted DRL framework with few additional parameters conforms to the notion of disentanglement and gets desirable results in practice. 

	%In the training process, the face encoder is usually pre-trained and frozen. The SFR training pipeline, with little  parameters to optimize, converges quite fast. 
	
	%In one case, this already meets the goal of decoupling and get desirable disentanglement results. Another reason is a face decoder introduce million parameters and it hardly converge in conventional SFR pipeline. Usually the face encoder is pre-trained, frozen and only the aggregation module is optimized in the training stage. For each training epoch, one identity is sampled rather than . 

	\begin{table*}[]
		\centering
		\begin{tabular}{l|c|c|c|c|c|c}
			\toprule
			\multirow{2}*{\textbf{Method}} & \multicolumn{3}{c|}{\textbf{IJB-B 1:1 TAR (\%)}} & \multicolumn{3}{c}{\textbf{IJB-C 1:1 TAR (\%)}}\\
			\cline{2-7} 
			~ & FAR=1e-6 & FAR=1e-5 & FAR=1e-4 &FAR=1e-6 & FAR=1e-5 & FAR=1e-4  \\
			\hline
			Vanilla (Entangled) & \st{41.51}  & 68.11 & 82.80&56.25&74.88&85.80\\
			+ Attention  & \st{38.31} & \textbf{72.94} & \textbf{86.10}&\textbf{ 66.17} &\textbf{80.08}&\textbf{ 88.61}  \\
			\hline
			Image-based disentangled & \st{40.08} & 70.49 & 83.81& 65.41 & 77.05 & 86.70  \\
			+ Attention (ID) &\st{42.04} & 69.67 & 83.42 & 60.97 & 75.60 & 86.07 \\
			+ Attention (VA) & \st{36.69} & \textbf{74.25} & \textbf{86.03}  &\textbf{71.87} & \textbf{81.08} & \textbf{88.58}\\
			\hline
			%\hdashline
			\textbf{Set-based disentangled} & \st{38.22} & 70.58 & 84.18 & 62.56 & 77.51 & 86.92\\
			+ Attention (ID) & \st{40.09} & 74.18 & 86.07&69.51 & 80.75 & 88.61\\
			\textbf{+ Attention (VA)} & \st{37.03} & \textbf{75.51} & \textbf{86.68} &\textbf{72.87} & \textbf{81.66} & \textbf{89.02}\\
			\bottomrule
		\end{tabular}
		% \vspace{-0.1in}
		\caption{Ablation of disentanglement and quality attention on IJB-C 1:1 verification protocols with VGGFace2 backbone. The best results are highlighted in bold.}
		\vspace{-0.4cm}
		\label{tab:disentangle}
	\end{table*}
	
	\subsection{Vocabulary-Based Aggregation} \label{sec:vagg}
	The proposed vocabulary-based aggregation (VBA) consists of variance-based quality assessment (VQA) and vocabulary-based burst suppression (VBS). 
	In VQA,  a vanilla attention block is applied on the disentangled variance features to get the quality scores as follow:
	\begin{equation}
	\alpha = \bm{\sigma} (q_{VA} f_{VA}^{T}).
	\end{equation}
	
	\noindent\textbf{Vocabulary-based burstiness suppression.}
	The disentangled variance features $f_{VA}$ are also utilized to indicate burstiness degree on the VBS method, which is based on the analysis on the reference-similarity matrix of $f_{VA}$ to a vocabulary $V \in \mathbb{R}^{k \times d}$.
	As illustrated in the right part of Figure~\ref{fig:pipeline}, the variance vocabulary consists of $k$ visual words, which is randomly initialized and end-to-end optimized in this paper.
	%or initialized by clustering algorithm K-means. 
	The variance features $f_{VA}$ are quantized by $V$ with matrix multiplication to get the assignment matrix $A$:
	\begin{equation}\label{eq:assign}
	A = f_{VA}^{} V^T, \ \ A \in \mathbb{R}^{n \times k}.
	\end{equation}
	
	To equalize the assignment scores of all the elements in the set, the inter-word normalization is first applied with softmax function to normalize the assignment score of each face. 
	\begin{equation}\label{eq:inter}
	A^{\prime} = Softmax(A\alpha, \ \ dim\!=\!2), 
	\end{equation}
	where $\alpha$ is a parameter to scale the numerical contrast.  $\alpha \to \infty $ corresponds to the hard-assignment situation, $\alpha > 0$ corresponds to soft-assignment and $\alpha = 0$ corresponds to equal-assignment. %We choose $\alpha = 10$ in practice.
	
	%The assignment scores of each face are normalized in the inter-normalization, now the summation of each face is equal to 1.
	
	%The assignment scores in Equation~\ref{eq:assign} are equal to the inner-product of features with the visual words, and the assignment score of a variance feature to its closest visual words are largest. 
	As formulated in Equation~\ref{eq:assign}, the assignment score of a variance feature to its closest visual words is largest. Thus the visual words with larger assignment norms are supposed to be bursty. To suppress the bursty faces and constrain the burstiness intra the words, an intra-word normalization operation is adopted to equalize the assignment norm of all the visual words. %are equalized and the bursty face features close to the bursty words are suppressed.
	Concretely, $L_1-$normalization is applied to the assignment scores intra each visual word, the bursty words with larger assignment norms are suppressed while the non-bursty words are relatively highlighted in this way.
	%we adopt intra-normalization to equalize all the visual words with $L_1$ normalization applied on each cluster and get 
	The normalized assignment matrix $A^{N}$ is formulated as:
	\begin{equation}\label{key}
	A^{N} = Intra(A^{\prime}) = L_1(A^{\prime}, \ \ dim\!=\!1). 
	\end{equation}
	
	%Because the assignment scores in each cluster are normalized in the intra-normalization, where scores in the bursty words are suppressed, and non-bursty are highlighted. A face close to the bursty words are expected to be bursty and suppressed, while the faces close to the non-bursty words are expected to be non-bursty and highlighted. 
	
	%Having the normalized assignment matrix where the bursty faces are suppressed, we then sum the assignment score of each face to its burstiness score. 
	
	Each row in $A^{N}$ corresponds to the normalized assignment scores of each face to all the words, where the scores assigned to bursty words are suppressed and the scores assigned to the non-bursty words are highlighted. Thus we sum the assignment scores of each element to the bursty scores $\beta$.  The bursty faces close to bursty words get lower bursty scores and the non-bursty get higher scores.
	%In this way, because all the words are normalized, the faces assigned to bursty words get lower assignment scores and the non-bursty get larger. We then sum the assignment weight of all the words to the bursty weights of each face. 
	\begin{equation} \label{key}
	\beta_{i} = \sum_{j} A_{i, j}^{N}. 
	\end{equation} 
	
	\noindent \textbf{Illustration of VBS}. To better explain the burstiness phenomenon and VBS, we take the hard assignment situation ($\alpha \to \infty $ in Equation~\ref{eq:inter}) as an example in Figure~\ref{fig:explain}. %,  where a face is assigned to one visual word.  
	Two feature sets $X = \{x_1, x_2, ..., x_{n}\}$ and $Y = \{y_1, y_2, ..., y_{n}\}$ are quantized by the vocabulary with four visual words. $V_1$ is regarded as a bursty word for $X$ with most features quantized, and $V_2$ is a bursty word for both $X$ and $Y$. 
	As is known, the similarity comparison of two sum-aggregated set representations corresponds to the summation of all-to-all cross-similarities of element features \cite{babenko2015aggregating, stylianou2019visualizing}. The cross-similarities as illustrated in Figure~\ref{fig:explain} (b) are dominated by the cross-similarities of bursty elements (\emph{i.e.} ${x_4, x_5, x_6}$ to ${y_2, y_3, y_4}$).

	When the intra-word normalization is applied to the assignment scores for each word, the elements in the bursty words get lower VBS weights (Figure~\ref{fig:explain} (c)). The cross-similarity with VBS weights multiplied is illustrated in Figure~\ref{fig:explain} (d), where all the words get equally similarity summations and none of the words dominate. That is to say, the set representations with VBS are democratic to all the words, namely all the variations.
	In practice, we adopt the soft-assignment ($\alpha = 10 $) to encounter quantization errors and get detailed assignment conditions \cite{liu2011defense}. %We will also scale the numerical ranges with softmax function. 
	
	%To suppress the burstiness phenomenon in a set, it is intuitive to apply existing burstiness suppression methods such as GMP or DA. But we found these methods never work and they have drawbacks in three aspects: \textbf{Firstly,} they are based on self-similarity matrix and thus are interfered by low-quality faces, which are distinct from other faces in a set and are regarded as unusual with larger weights. So we found they even damage the performance in the unconstrained scenario. \textbf{Secondly,} GMP and DA requires the Gram matrix and the computation complexity is $O(n)$, Moreover, GMP requires to calculate the matrix inverse, and DA iteratively with sinkhorn algorithm. The propposed VB with complexity is $O(nk)$ is applicable and Scalable to sets of any sizes.  %Thirdly, They requires to re-caluculate the Gram matrix when adding the sets. 
	
	%To remedy these drawbacks, we propose a vocabulary-based aggregation with disentangled variance vocabulary. We propose the reference similarity (assignment scores) rather than self-similarity to indicate the burstiness degrees.with computation complexity is $O(nk)$, where k is the cardinality of a referenced vocabulary. The benefit are threefold, reference-based similarity is not interfered by low-quality faces. Second, computation simple . Thirdly, Scalable.
	
	Having the disentangled identity features $f_{ID}$ discriminative for identity recognition, the VQA quality scores $\alpha$ and VBS burst scores $\beta$ assessed on the disentangled variances features, we apply weighted sum aggregation to get the final set representation $F$.  
	\begin{equation} \label{eq:f}
	F = \sum\nolimits_{i=1}^{n} \alpha_{i} \beta_{i} f_{ID}^i, 
	\end{equation}
	
	\noindent\textbf{Training objectives.}
	In each training epoch, all the identities are sampled and each identity corresponds to one training instance (set), which consists of $n_t$ randomly sampled faces of same identity. The identity features and variance features in an instance are swapped with each other in the disentanglement process ($L_{recons-img}$), and the set representation $F$ is also adopted to replace the original identity features for reconstruction ($L_{recons-set}$).
	The set representation $F$ is used for identity classification with cross-entropy loss $L_{CE}$. 
	The total training objective is formulated as:
	\begin{equation}
	L = L_{CE} +  L_{recons-img} +  L_{recons-set}
	\end{equation}

	%The set representations are thus discriminative and democratic to all the variations in the set. In the training stage, $F$ is optimized for identity classification with CE loss, and the whole training objective together with disentanglement loss is:

	\iffalse
	\noindent\textbf{Compared to VLAD}
	As described above the proposed VBA use analysis on the assignment matrix to get the burstiness scores. 
	VLAD is also known as a vocabulary-based aggregation method which use the assignment matrix to aggregate the element features. 
	
	But we argue that VLAD is not suitable for set-based FR for two reasons: Firstly, VLAD selectively quantize heterogeneous features (i.e., local features in image) to different cluster and concatenate the cluster features to representation. The faces in set, however, are homogeneous. Two faces quantized to different cluster get  VLAD representation 
	
	homogeneous
	different from the aggregation of local feature with different classes, the faces in set are homogeneous of same identity. So the quantization of faces of same identity requires disentangled variance vocabulary. 
	
	So, compared to VLAD, the proposed V-aggregation is advanced in rather than concatenate. 
	
	In our practice, VLAD lose its selectivity and degrade to global average pooling when faces in set are usually assigned to same cluster, and fail to when faces in set are usually assigned to same cluster. What is more, we found training VLAD representation on VGGFace usually cause , and . 
	\fi
	
	\section{Experiments} \label{sec:experiment}
	%1. CP-IJB-C and CQ-IJB-C Datasets. Demodalizing Face Recognition with Synthetic Samples
	
	\subsection{Datasets}
	We adopt the VGGFace2 \cite{cao2018vggface2} (SENet-50 \cite{hu2018squeeze}) or the ArcFace \cite{deng2019arcface} (ResNet-101 \cite{he2016deep}) model as the face encoder, which is separately pre-trained on the VGGFace2 \cite{cao2018vggface2} or a cleaned version  of MS1M \cite{guo2016ms} dataset. The whole disentanglement framework is trained for set-based face recognition on the VGGFace2 dataset.
	
	The unconstrained face recognition benchmarks, IJB-B \cite{whitelam2017iarpa} and IJB-C \cite{maze2018iarpa}, are used for evaluation. IJB-B and IJB-C are mix-sourced (video frames and images) benchmarks for set-based face verification (1:1) and identification (1:N). 
	We also have experiments on the Youtube Face (YTF) dataset \cite{wolf2011face} which is video-sourced and the set-based verification performance is evaluated. 
	
	\subsection{Implementation Details}
	%There are two . First adopt global averaging pooling to get face features, the backbone is trained on VGGFace2 or private data. Second is the recently on MS1M, and use a fully connected layer. Second is more because more identity and more parameters introduced by the FC layer and recently more liuxing.
	
	The MTCNN algorithm \cite{zhang2016joint} is adopted for face detection and alignment. Following previous works \cite{cao2018vggface2, deng2019arcface}, the faces are detected, cropped, without alignment, and resized to $224 \times 224 \times 3$ as inputs of VGGFace2 model. The faces are detected, cropped, aligned, and resized to $112 \times 112 \times 3$ for ArcFace model. 
	%Following the face encoder, we apply a dropout layer with dropout ratio 0.1 before aggregation.
	
	The VGGFace2 face encoder outputs 2,048-dimensional features,  which are projected to 256-dimensional identity features and variance features in the disentanglement encoder. The ArcFace features are 512-dimensional and are projected to 512-dimensional identity features and variance features in the disentanglement encoder.
	The vocabulary consists of 32 randomly initialized visual words for both VGGFace2 and ArcFace models.
	%The feature dim, VA, ID. Vocab size, random initialize.
	In the training stage, a training instance consists of 15 faces randomly sampled from a corresponding set. The parameters of face encoder are frozen and other modules are optimized with the RMSprop optimizer, batch size of 128, learning rate of 0.001, momentum of 0.9. %We set the parameters in Section~\ref{sec:method} $\lambda = 1$, $\lambda_1 = 0.5$. %margin m = 0.4
	The codes will be released after the paper is accepted.

	\begin{table*}[]
		\centering
		\begin{tabular}{l|c|c|c|c|c|c}
			\toprule
			\multirow{2}*{\textbf{Method}} & \multicolumn{3}{c|}{\textbf{IJB-B 1:1 TAR (\%)}} & \multicolumn{3}{c}{\textbf{IJB-C 1:1 TAR (\%)}}\\
			\cline{2-7} 
			~ & FAR=1e-6 & FAR=1e-5 & FAR=1e-4 &FAR=1e-6 & FAR=1e-5 & FAR=1e-4  \\
			\hline
			Vanilla (Entangled) & \st{41.51}  & 68.11 & 82.80&56.25&74.88&85.80\\
			%+ Attention  & \st{38.31} & 72.94 & 86.10& 66.17 &80.08& 88.61  \\
			+ VBS  & \st{39.97} &\textbf{75.09} & \textbf{86.20}&  69.80 & \textbf{80.83} & \textbf{88.64}\\
			+ VBS + Attention & \st{37.20}& 74.27 & 85.89&\textbf{69.89} & 80.50 & 88.46\\ 
			\hline
			\textbf{Set-based disentangled} & \st{38.22} & 70.58 & 84.18 & 62.56 & 77.51 & 86.92\\
			+ VBS (ID) &  \st{39.85} & 69.49 & 83.78& 61.36 & 76.94 & 86.87\\ 
			\textbf{+ VBS (VA)} &  \st{36.72} & 76.01 & 86.27&72.67 & 80.61 & 88.74\\ %38.65 & 71.80 & 84.76& 64.54 & 78.44 & 87.54
			\textbf{+ VBS (VA) + VQA} & \st{35.76} & \textbf{76.83} & \textbf{86.83} & \textbf{75.04} & \textbf{82.18} & \textbf{89.14} \\
			\bottomrule
		\end{tabular}
		% \vspace{-0.1in}
		\caption{Ablation of vocabulary-based burst suppression on entangled and disentangled models with VGGFace2 backbone. }
		%\vspace{-0.2cm}
		\label{tab:disentangle_burst}
	\end{table*}
	\subsection{Ablation Study}
	In this subsection, we give exhaustive ablation studies to demonstrate the effectiveness of set-based disentanglement (SD) framework, variance-based quality assessment (VQA) and vocabulary-based burstiness suppression (VBS).
	All the methods in this subsection are end-to-end optimized unless otherwise noted. %Most of the ablation is based on the VGG because it is a weak model and the gaps are prominent.
	
	%\noindent\textbf{The impact of disentanglement and variance-based quality attention.} 
	\subsubsection{The Impact of Disentanglement and VQA} 
	
	To demonstrate the effectiveness of the proposed light-weighted SD framework, we compare three kinds of features: the vanilla entangled features, image-based disentangled identity features (Equation~\ref{eq:imgd}) and the proposed SD identity features ((Equation~\ref{eq:imgd} $+$ Equation~\ref{eq:setd})). Then we apply quality attention to these features and give the result comparisons in Table~\ref{tab:disentangle}. 
	We found the results of the 1:1 TAR when FAR=1e-6 are not stable on IJB-B dataset and suppose the reason is there are less (8 million) unmatched positive pairs. So we apply strikethrough on these results and recommend the reader to focus on the other metrics.
	
	The first observation that can be drawn is that disentanglement improves the feature robustness, and set-based disentangled features usually outperform entangled features and image-based features. 
	We can also observe the attention block consistently improves the recognition performance for three features. The attention blocks based on variance features are usually more prominent than identity-based and the entangled-based blocks, which demonstrates the effect of variance-based quality assessment (VQA).
	
	%We first evaluate their recognition performance, then apply vanilla attention-aware aggregation on them and the performance comparison is shown in Table~\ref{label}.  the . As can be seen, the disentangled ID features are rubust to variance, thus are most discriminative than entangled or the variance features. While the attention module on ID features even decrease the performance because ID itself doesn't indicate face quality, while Variance do. We can observe attention on variance features get best performance than the entangled and the ID. The proposed VB attention further improve the performance.
	
	%\textbf{We can also observe, The first observation can be drawn is that }
	
	%The disentangled ID features are more discriminative than the entangled because the variance features are separated by disentanglement. We can also observe that the VA-based attention surpass ID-based attention and the entangled attention, because face quality is variance-related and variance features are pure to indicate face quality. Similarly, we can observe VB aggregation based on VA vocabulary get better results than entangled and ID.
	
	%\noindent\textbf{The impact of variance-based burst assessment.} 
	\subsubsection{The Impact of Variance-Based Burst Suppression.} 
	
	To demonstrate the effectiveness of VBS, we apply it to the vanilla entangled model and SD model.  As can be seen in Table~\ref{tab:disentangle_burst}, the proposed VBS is effective on both the entangled and disentangled models with considerable performance improvement. 
	For the SD model, the VBS on identity features gets worse results because the face burstiness is identity-unrelated. 
	In addition, VBS is complementary to the attention mechanisms VQA on SD model, and their combination further improves the performance. 
	Note that the VGGFace2 model is adopted for the ablations in Table~\ref{tab:disentangle} and Table~\ref{tab:disentangle_burst}. The observations on the ArcFace-based model are similar and corresponding result comparisons are given in the supplementary.
	
	%To demonstrate the effectiveness of variance features to the assessment of face quality, we apply vanilla attention module (Equation~\ref{label}) on entangled features, identity (ID) features and variance features (VA). To demonstrate the effectiveness of variance features to the assessment of face burstiness, we apply VB burst assessment on entangled features, identity (ID) features and variance features (VA). 
	
	%\noindent\textbf{Complementary to existing attention mechanisms.} 
	\subsubsection{Complementary to Elaborate Attention Mechanisms.} 
	
	As observed in Table~\ref{tab:disentangle_burst}, the proposed set-based disentanglement (SD) and vocabulary-based burst suppression (VBS) are complementary to vanilla attention module. We also combine them with two elaborate attention modules, the NAN \cite{yang2017neural} and MCN \cite{xie2018multicolumn}. We follow the implementation details of the original papers, and give the result comparisons in Table~\ref{tab:disentangle_att}. 
	As can be seen, three attention blocks get evenly matched results with the original entangled features. When combined with SD and VBS, it is interesting to find that vanilla attention block performs slightly better. 
	
	\begin{table}[]
		\centering
		\begin{tabular}{l|c|c|c|c|c|c}
			\toprule
			\multirow{2}*{\textbf{Method}} & \multicolumn{3}{c|}{\textbf{IJB-B 1:1 TAR (\%)}} & \multicolumn{3}{c}{\textbf{IJB-C 1:1 TAR (\%)}}\\
			\cline{2-7} 
			~ & 1e-6 & 1e-5 & 1e-4 & 1e-6 & 1e-5 & 1e-4 \\
			\hline
			%&\multicolumn{3}{c|}{\textbf{VGGFace2}} &\multicolumn{3}{c}{\textbf{MS1M}}\\
			NAN \cite{yang2017neural} & \st{36.96} & 72.86 & 85.70&69.01 & 80.15 & 88.29\\
			\textbf{+ SD + VBS} &\st{34.54} & \textbf{76.03} & \textbf{86.59}&\textbf{71.81} & \textbf{81.59} & \textbf{89.05}\\
			\hline
			MCN \cite{xie2018multicolumn}& \st{36.25} & 72.52 & 85.83&68.53& 79.94& 88.21\\
			\textbf{+ SD + VBS}  & \st{35.30} & \textbf{76.95} & \textbf{86.47}& \textbf{74.56} & \textbf{82.34} & \textbf{89.04}\\
			\hline
			Vanilla &\st{38.31} & 72.94 & 86.10& 66.17 &80.08& 88.61\\
			\textbf{+ SD + VBS} &\st{35.76} & \textbf{76.83} & \textbf{86.83} & \textbf{75.04} & \textbf{82.18}& \textbf{89.14}\\
			\bottomrule
		\end{tabular}
		% \vspace{-0.1in}
		\caption{Ablation on the combination of the proposed SD, VBS and three attention blocks with VGGFace2 backbone.}
		%\vspace{-0.2cm}
		\label{tab:disentangle_att}
	\end{table}
	\begin{table}[]
		\centering
		\begin{tabular}{l|c|c|c|c|c}
			\toprule
			\multirow{2}*{\textbf{Vocabulary}} & \multicolumn{3}{c|}{\textbf{IJB-C 1:1 TAR (\%)}} & \multicolumn{2}{c}{\textbf{YTF}}\\
			\cline{2-6} 
			~ & 1e-6 & 1e-5 & Time (s) & Acc. & Time (s)\\
			\hline
			&\multicolumn{5}{c}{\textbf{VGGFace2}} \\
			SD model &63.57 & 77.69 &85.36 &94.88& 11.87\\   %
			+ GMP \cite{murray2014generalized}&62.84 & 77.43 & 166.75&95.06&24.42\\ %
			+ DA \cite{jegou2014triangulation}& 63.24 & 77.63 & 275.64 &95.22&31.57\\  %30.30
			\textbf{+ VBS} &  \textbf{72.67} & 80.61 & \textbf{95.17}&95.08&\textbf{12.32}\\    %
			\hdashline
			\textbf{+ VBS + GMP} & 72.49 &\textbf{80.63} &175.36 &95.30&25.48\\
			\textbf{+ VBS + DA} & 71.93 & 80.49  & 278.31& \textbf{95.54}&33.34\\
			%VB \\
			\hline
			&\multicolumn{5}{c}{\textbf{ArcFace}} \\
			SD model & 86.21 & 92.72 &80.67&96.70&9.05\\  %38.28 | 89.33 | 94.25
			+ GMP \cite{murray2014generalized}& 86.32 & 92.58 &147.75 &96.98&20.51\\
			+ DA \cite{jegou2014triangulation} & 85.92 & 92.70 & 245.34&97.00&32.28\\   
			\textbf{+ VBS} &89.06 & \textbf{93.90} & \textbf{94.28}&96.96&\textbf{9.39}\\  %86.49 | 93.06 | 95.46   %97.26 GMP+VB
			\hdashline
			\textbf{+ VBS + GMP} &\textbf{89.19}& 93.74 &161.48&\textbf{97.26}&22.25\\
			\textbf{+ VBS + DA} & 86.86 & 93.38 & 256.44&97.14&34.66\\
			%VB  \\ 
			\bottomrule
		\end{tabular}
		\caption{Ablation of three burstiness suppression methods on the  set-based disentanglement model.}
		\vspace{-0.2cm}
		\label{tab:burst_supp}
	\end{table}
	\begin{table*}[t]
		\centering
		\begin{tabular}{l|c|c|c|c|c|c|c}
			\toprule
			\multirow{2}*{\textbf{Method}} &\multirow{2}*{Backbone} & \multicolumn{3}{c|}{\textbf{IJB-B 1:1 TAR (\%)}} & \multicolumn{3}{c}{\textbf{IJB-C 1:1 TAR (\%)}} \\
			\cline{3-8} 
			~ &&FAR=1e-6& FAR=1e-5 & FAR=1e-4 & FAR=1e-6& FAR=1e-5 & FAR=1e-4 \\
			\hline
			% VGGFace2 & VGGFace2& 64.7  & 78.4 &  \\
			VGGFace2 \cite{cao2018vggface2} &\multirow{7}*{VGGFace2}& -&67.1 & 80.0 & 61.7 &75.3&85.2\\ 
			MCN \cite{xie2018multicolumn}& && 70.8 & 83.1 & 64.9 &77.5&86.7  \\  
			%DCN \cite{xie2018comparator} & &-&- & 84.9 & -&- & 88.5 \\
			GhostVLAD \cite{zhong2018ghostvlad} && - &76.2 & 86.3 &-&-&-   \\
			PCNet \cite{xie2020inducing} & &- &- &- &69.5 &80.0&89.0 \\
			\textbf{SD + VBS}& &\st{36.72} & 76.01 & 86.27&72.67 & 80.61 & 88.74 \\
			\textbf{SD + VBA}& &\st{35.76} & \textbf{76.83} & \textbf{86.83} & \textbf{75.04} & \textbf{82.18} & \textbf{89.14} \\
			\hline
			%PFE \cite{shi2019probabilistic} & \multirow{5}*{ArcFace} &-&-&-&-& 89.64& 93.25 \\ %IJBC	
			%ArcFace \cite{deng2019arcface} & \multirow{2}*{MS1M-C}& 90.82&94.90&84.00&94.16&95.14& 96.87\\   
			ArcFace \cite{deng2019arcface} & \multirow{3}*{ArcFace}&\st{38.28}& 89.33&94.25& 86.25&93.15&95.65\\   	
			%\textbf{Attention}& &\textbf{} & \textbf{} &  \\
			\textbf{SD + VBS}&&\st{46.39} & 90.45 & 94.34&89.06 & \textbf{93.90} & 95.84   \\
			%\textbf{SD + VBA} & & \st{44.03} &90.41& 94.22 & 89.25 & 93.77 & 95.80\\ ALPHA=10
			\textbf{SD + VBA} & & \st{43.49} & \textbf{90.72} & \textbf{94.38}&\textbf{89.31} & 93.82 & \textbf{95.85}\\ %ALPHA=20
			\hline
			AdaFace \cite{kim2022adaface} & \multirow{3}*{AdaFace}&\st{48.19}& 89.25 &\textbf{95.42}&87.90&94.22 &96.63\\   	
			%\textbf{Attention}& &\textbf{} & \textbf{} &  \\
			\textbf{SD + VBS}&&\st{47.71}&90.66&95.39& 89.74&94.76&\textbf{96.65}   \\
			%\textbf{SD + VBA} & & \st{44.03} &90.41& 94.22 & 89.25 & 93.77 & 95.80\\ ALPHA=10
			\textbf{SD + VBA} & & \st{48.96} & \textbf{90.89}&95.38&\textbf{90.02}&\textbf{94.85}&\textbf{96.65}\\ %ALPHA=20
			%\textbf{SD + VBA + Norm} & &43.29 & 90.84 & 94.41  & 90.30 & 94.09 & 95.85 \\ %ALPHA=20
			%\textbf{SD + VBS + Norm} &&\st{46.16} & \textbf{90.92} & \textbf{94.41}&\textbf{90.90} & \textbf{94.18} & \textbf{95.87}\\
			\bottomrule
		\end{tabular}
		% \vspace{-0.1in}
		\caption{Compared with state-of-the-art set-based face recognition works on IJB-B and IJB-C 1:1 verification protocols. }
		\label{tab:SOTA}
		\vspace{-0.2cm}
	\end{table*}
	
	%\noindent\textbf{Compared to existing burstiness suppression methods.} 
	\subsubsection{Compared to Existing Burstiness Suppression Methods.} 
	%Note that VB aggregation is, it act as quantizer to quantize groups. The groups of are given in IJB dataset, but these groups are not always available. The proposed can act group generator. Different from Q-shift, it is applicable to set with any sizes. We found it perform well when the groups are unknown, and are also act when the groups are known. 
	
	GMP \cite{murray2014generalized} and DA \cite{jegou2014triangulation} are two representative burstiness suppression methods. Even though they consist of  differentiable operations, we found they get unsatisfactory results when trained end-to-end.  So we apply them as a post-processing method on the proposed SD model (without VQA and VBS), and have comparison with VBS on the IJB-C and YTF dataset. IJB-C is an unconstrained benchmark with some amount of low-quality faces. YTF is a video-based face recognition benchmark with small amount of low-quality faces, and the face quality in a video does not change much.
	
	We give the comparison of the proposed VBS to GMP and DA in Table~\ref{tab:burst_supp} on two datasets and two face encoders.
	As can be seen, the observations are quite similar for both the VGGFace2 and the ArcFace model, while the observations on two datasets are different.
	GMP and DA hardly take effect on the IJB-C dataset, because they are interfered with by the low-quality faces in the unconstrained scenario. 
	When the interference gets lower, GMP and DA take effect on the YTF dataset and slightly outperform VBS with both backbone. Comparatively, the proposed VBS consistently improves the SD model on both backbones and both datasets. 
	We also combine VBS with GMP and DA, the performance on YTF dataset is further improved while there are few changes on IJB-C dataset because of the interference of low-quality faces.
	
	From these observations, we can summarize that the proposed VBS with analysis on reference-similarity matrix is robust to low-quality faces and effective on both the unconstrained and constrained scenario. The self-similarity-based burst suppression methods, GMP and DA, are interfered with by low-quality faces and only applicative on the somewhat constrained scenario, where they are complementary to the proposed VBS.
	
	The computation cost of three burstiness suppression methods are also presented in Table~\ref{tab:burst_supp}, under the environment of 48 cores CPU, single RTX 1080Ti GPU, CUDA
	10.1, Torch framework. Only the aggregation time (averaged on 5 times) of the whole dataset is reported because it is brief compared to the feature extraction time. 
	%The IJB-C dataset consists of 23,124 face sets with average of 20.3 faces. YTF dataset consists of 3425 videos with average of 181.3 face frames. %As aforementioned, the computation complexity of gram matrix and assignment matrix is separately $\mathcal{O}(n^2)$ and $\mathcal{O}(nk)$. 
	As can be seen, the proposed VBS brings little additional computation cost on the SD model. The computation costs of GMP and DA on both datasets and backbones are comparatively larger. %because GMP and DA separately inverse the regularized gram matrix and adopt a modified Sinkhorn algorithm to get the burst weights.
	%The DA iteratively computes the burst weights with a modified Sinkhorn algorithm and GMP gets the closed-form solution of the linear-regression problem by inversing the regularized gram matrix. 

	%Especially in the YTF dataset with large set cardinality. 
	%In this way, the propsoed VBA is an effective and efficient method, and it is also complementary to GMP on the video-based SFR benchmark.
	
	%In this regard the proposed VB can be trained end-to-end, and it can also be implemented as an unsupervised way where the vocabulary is initialized by an clustering algorithm. For fair comparison, we apply GMP, DA, unsupervised VBA as the post-processing methods on the original VGGFace2 and ArcFace features, and show the result comparison in Table~. It can be seen that . 
	\begin{table}[]
		\centering
		\begin{tabular}{l|c|c|c|c|c|c}
			\toprule
			\multirow{2}*{\textbf{SD + VBA}} & \multicolumn{3}{c|}{\textbf{IJB-B 1:1 TAR (\%)}} & \multicolumn{3}{c}{\textbf{IJB-C 1:1 TAR (\%)}}\\
			\cline{2-7} 
			~ & 1e-6 & 1e-5 & 1e-4 & 1e-6 & 1e-5 & 1e-4 \\
			\hline
			%P-norm \\
			$k=8$ & \st{38.95} & 75.60 & 86.42&72.38 & 81.08 & 89.01\\%
			$k=16$ & \st{35.45} & 75.43 & 86.27&73.27 & 81.06 & 88.78\\%
			$k=32$ & \st{35.76} & \textbf{76.83} & \textbf{86.83} & \textbf{75.04} & \textbf{82.18} & \textbf{89.14}\\%
			$k=64$ & \st{38.93} & 76.80 & 86.50&73.18 & 81.81 & 89.07\\%36.27 | 76.18 | 86.26
			\hline
			$\alpha=1$ & \st{39.68} & 74.26 &86.33&72.47 & 81.38 & 88.90 \\% 
			$\alpha=5$ & \st{37.28} & \textbf{77.05} & \textbf{86.83}&73.72 & 82.11& \textbf{89.28}\\%
			$\alpha=10$ & \st{35.76} & 76.83 & \textbf{86.83} & \textbf{75.04} & \textbf{82.18}& 89.14\\%
			$\alpha=20$ & \st{37.66} & 75.72 & 86.09 & 73.16 & 80.95& 88.52\\% 
			\bottomrule
		\end{tabular}
		\caption{Ablation of parameters in vocabulary-based burst suppression with VGGFace2 backbone.}
		\vspace{-0.4cm}
		\label{tab:ablation_vba}
	\end{table}
	\subsubsection{Ablation of Vocabulary-Based Burst Suppression.} 
	
	We give ablations on the parameters of VBS on different choices of vocabulary size $k$ and scale $\alpha$ in inter-normalization. The SD model with vocabulary-based aggregation (VBA with VBS and VQA) is adopted and the ablations are shown in Table~\ref{tab:ablation_vba}. An 8-word VBS gets considerable results on both datasets and usually best results are obtained when $k = 32$. As shown in the bottom part, the VBS usually gets consistently considerable results when $\alpha = 10$. We thus choose $k = 32$ and $\alpha = 10$ for the VBS method by default.
	
	%Note that after training, we extract VA features from sampled training set and adopt sphere K-means to recalculate the vocabulary. Because the vocabulary optimized by end-to-end learning is not accurate than the vocabulary got by the clustering algorithm. We also investigate the performance with varying vocabulary size. 8, 16, 32, 64.
	\begin{table}[]
		\centering
		\begin{tabular}{l|c|l|c}
			\toprule
			\textbf{Method}&Acc. (\%) & \textbf{Method} & Acc. (\%) \\
			\hline
			Eigen-PEP \cite{li2014eigen} & 84.8& DeepFace \cite{taigman2014deepface}& 91.4\\ 
			FaceNet \cite{schroff2015facenet}& 95.52 & DAN \cite{rao2017learning}& 94.28 \\
			DeepID2+ \cite{sun2015deeply} & 93.20 & QAN \cite{liu2017quality}&96.17\\
			%NAN \cite{yang2017neural} & 95.72 & REAN \cite{gong2019recurrent} & 96.60\\
			%Liu \emph{et al.} \cite{liu2019feature} & 96.21 & C-FAN \cite{gong2019video}& 96.50\\ 
			NAN \cite{yang2017neural} & 95.72 & C-FAN \cite{gong2019video}& 96.50\\ 
			CosFace \cite{wang2018cosface} & 97.65 & ArcFace \cite{deng2019arcface}& 98.02\\ 
			\hline
			ArcFace-C \cite{deng2019arcface} & 96.66  &\textbf{SD + VBA}& 96.94\\%
			%+ SD + VBA + VQA& 96.76\\%97.
			\textbf{SD + VBA + GMP}& 96.96&	\textbf{SD + VBA + DA}& \textbf{97.22}\\%97.26
			%\textbf{SD + VBA + DA}& \textbf{97.22}\\%97.26
			\bottomrule
		\end{tabular}
		\caption{Video face verification performance on YTF dataset.}
		\vspace{-0.3cm}
		\label{tab:YTF}
	\end{table}
	\begin{figure*}[t]
		\centering
		\begin{subfigure}[t]{0.5\textwidth}
			\centering
			\includegraphics[scale = 0.43]{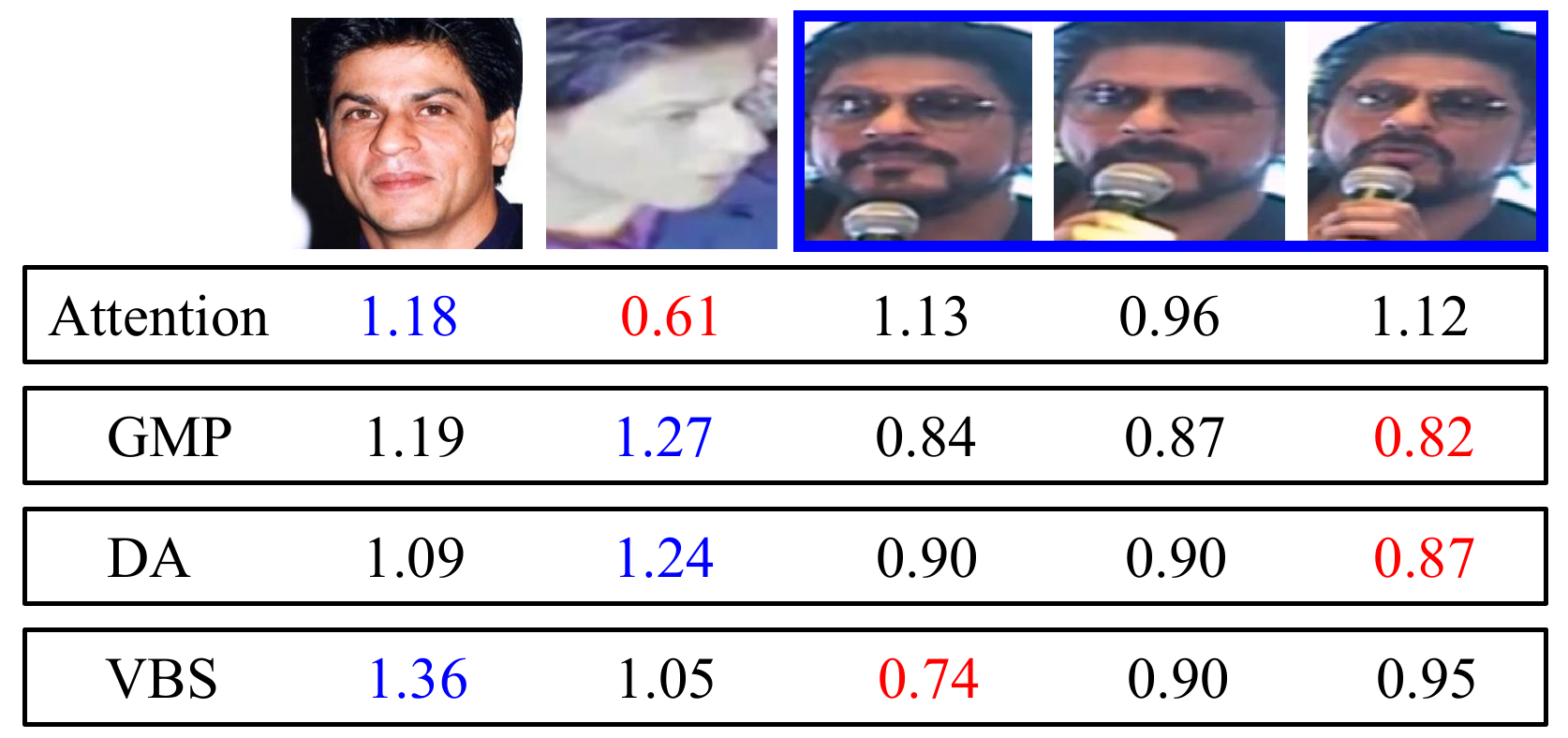}
			%\captionsetup{labelformat=empty}
			%\caption{Regular-size models}
		\end{subfigure}%
		\begin{subfigure}[t]{0.5\textwidth}
			\centering
			\includegraphics[scale = 0.43]{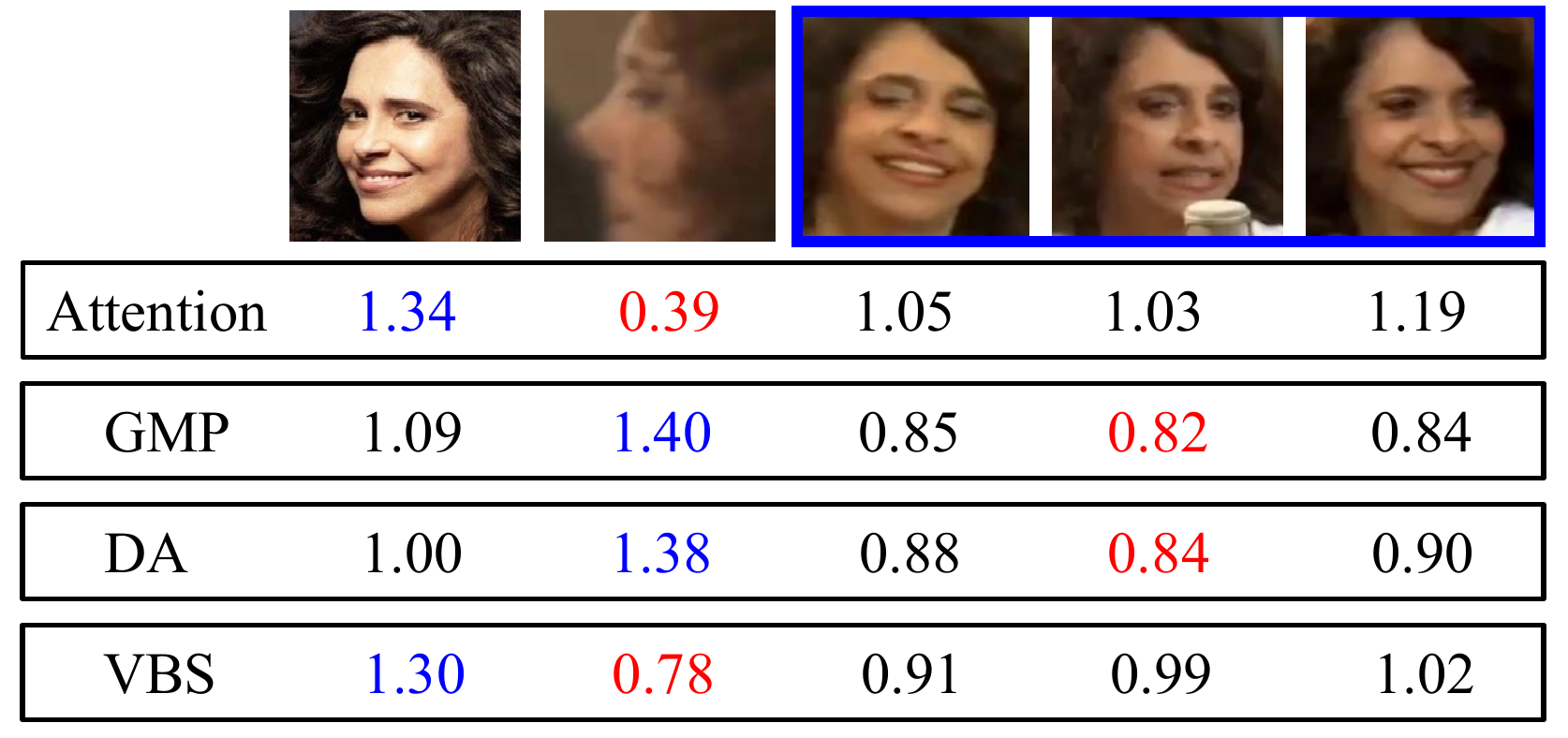}
			%\captionsetup{labelformat=empty}
			%\caption{Mobile-size models}
		\end{subfigure}%
		\caption{Visualization of the estimated quality attention, GMP, DA and VBS scores in the set. The scores are normalized for comparison.  The highest and lowest values are highlighted in blue and red. }
		\label{fig:burst_score}
	\end{figure*}
	\begin{figure*}[t]
		\begin{center}
			\includegraphics[scale=0.44]{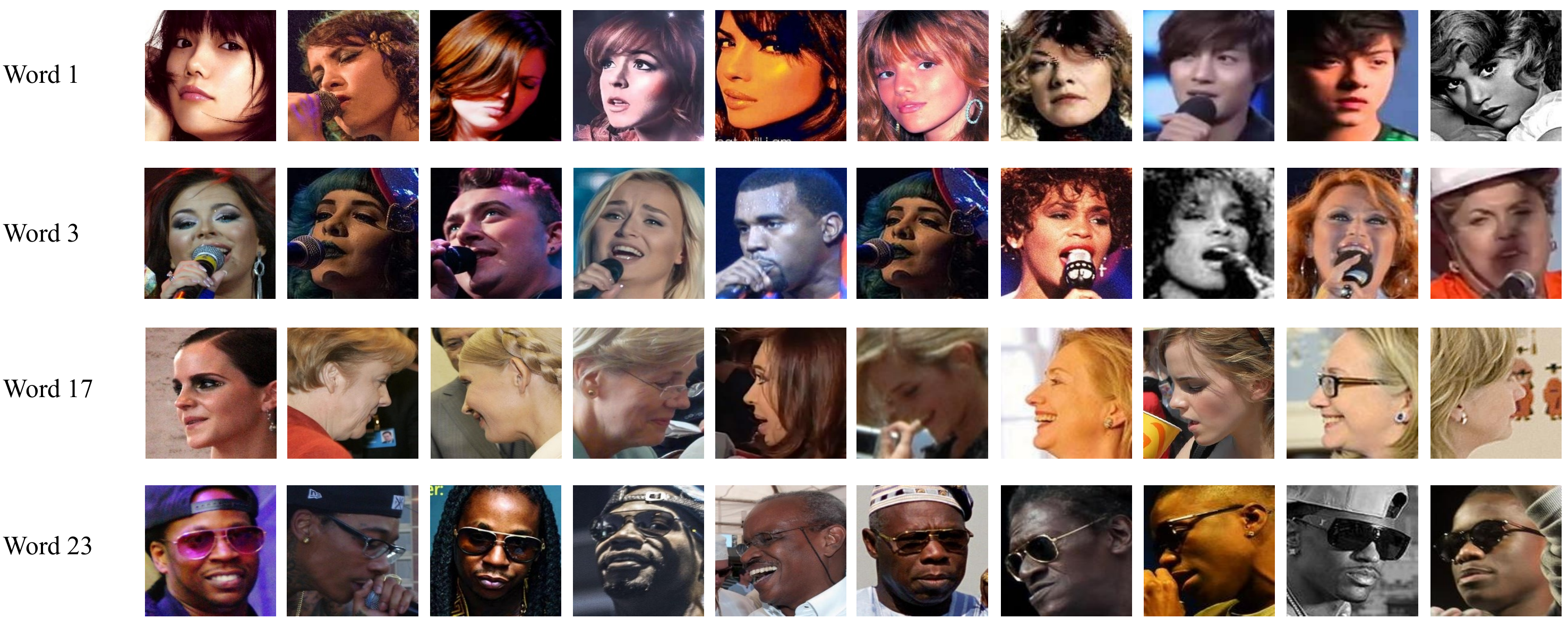}\vspace{-0.2cm}
		\end{center}
		% \vspace{6pt}
		\caption{Examples of the representative faces quantized by the trained visual words.
		}
		\label{fig:cluster_vis}
		%\vspace{-0.1cm}
	\end{figure*}
	\subsection{Comparison with State-of-the-Arts} \label{sec:sota}
	
	We compare with the state-of-the-art works on IJB-B and IJB-C datasets in Table~\ref{tab:SOTA}. On the VGGFace2 backbone, the attention mechanisms \cite{yang2017neural, xie2018multicolumn, xie2020inducing} are widely studied. %and hard-pair mining \cite{xie2018comparator} strategy is also proposed. 
	The GhostVLAD \cite{zhong2018ghostvlad} adopts VLAD aggregation with a ghost cluster to eliminate the low-quality faces.
	The proposed method is based on the set-based disentanglement (SD) framework and vocabulary-based aggregation (VBA), which consists of vocabulary-based burstiness suppression (VBS) and variance-based quality assessment (VQA). 
	
	It can be seen that the VBS is complementary with VQA in VBA to improve the performance of the proposed SD framework. The SD with VBA considerably surpasses the state-of-the-art works on both datasets with the VGGFace2, ArcFace and AdaFace \cite{kim2022adaface} backbone, especially on the IJB-C dataset. 
	%For the ArcFace model, we additionally replace the VQA with the original feature norms as quality scores. It is interesting to observe that the original feature norm is more representative to face quality than VQA.
	Note that few of existing works report the 1:N identification performance, so we present the 1:N identification results in the supplementary.
	
	\noindent \textbf{Youtube Face.}  For the evaluation of YTF dataset, most of existing works adopt exhaustive (all-to-all) matching on the verification of set pairs and the maximal matching score is used as the pair similarity. This practice brings large storage and computing overhead, which is not scalable to large-scale recognition. So we adopt sum-aggregation and re-evaluate the ArcFace model on YTF with compact set representations (ArcFace-C). 
	We then apply the proposed method SD framework with VBA on the ArcFace backbone. The results are shown in Table~\ref{tab:YTF}, where SD and VBA improve the vanilla ArcFace model. The proposed method is complementary to the self-similarity-based burst suppression methods, GMP and DA, to further improve the performance.

	\subsection{Qualitative Results} 
	As aforementioned,  the existing burstiness suppression methods, GMP and DA, %are based on the analysis of self-similarity matrix and 
	are interfered with by low-quality faces. To verify this argument, we give visualizations of the estimated quality attention scores, GMP scores, DA scores and the proposed VBS scores in Figure~\ref{fig:burst_score}. 
	The second faces in both examples with lowest quality get highest GMP and DA scores, because they are most unusual in the set. But the low-quality face features are less discriminative and highlighting them damages the set representation. 
	For the VBS scores, the frequent faces are suppressed with lower weights and the face with high quality and low frequency usually gets highest scores. Even though the low-quality faces are also highlighted, the extent is slighter than GMP and DA. %We think this phenomenon is inevitable and acceptable because VBS practically improves the performance on unconstrained scenario. 
	
	%relatively robust to low-quality and give highest scores to the unfrequen faces with high discriminability.
	
	%The proposed VBA based on analysis of reference-similarity matrix gives lower to the frequent faces, and slightly . Even though VBA also improve the low-quality faces, it is slight and acceptable when simultaneously suppress the bursty faces.

	To reveal what is learned in the vocabulary $V$ with $k$ visual words, we randomly sample 
	variance features of 20,000 faces in IJB-C dataset and compute the reference-similarity matrix from them to the trained vocabulary with 32 visual words. 
	The representative faces closest to some words are illustrated in Figure~\ref{fig:cluster_vis}.  It can be seen that the faces assigned in the words are identity-unrelated and variance-related.
	The faces assigned in word 1 are usually with similar hair styles, and the faces in word 3 are  with microphones. The faces quantized by word 17 are usually female profiles and the faces quantized by word 23 are usually black men with sunglasses.

	\section{Conclusion}
	%In this paper, we give an exhaustive study on the burstiness phenomenon in set-based face recognition task. We propose three strategies to detect the bursty faces in sets and separately apply them in the training stage with burst-aware instance sampling and in the evaluation stage with burst-aware aggregation. Moreover, we propose the QA-GMP to combine GMP with face quality attention, which enables the original GMP to be robust to the low-quality face features. Extensive experiments demonstrate the effectiveness of burstiness suppression with proposed strategies, which get new state-of-the-art results on the SFR benchmarks.
	
	In this paper, we propose a light-weighted set-based disentanglement (SD) framework with vocabulary-based aggregation (VBA) for discriminative and representative set representations. The disentangled identity features are discriminative to variances and the disentangled variance features are fully utilized in the proposed vocabulary-based burst suppression (VBS) and variance-based quality assessment (VQA) methods  to indicate face quality and burstiness. The proposed VBS is the first method to suppress the burstiness problem for SFR task, and is applicable on various scenarios  with prominent effectiveness and efficiency.
	%Extensive illustrations explains the underlying mechanisms of the burstiness problem and the VBS method.
	Extensive experiments demonstrate the effectiveness of disentanglement with VBA, which gets new state-of-the-art results on the SFR benchmarks.
	%to separate the identity features with the variance features in a light-weighted set-based disentanglement framework. 

\section*{Acknowledgments}
This work was supported in part by the Zhejiang Natural Science Foundation under Grant LR19F020006, National Natural Science Foundation of China under Grant No.61836002, National Natural Science Foundation of China under Grant No. 62072397, National Key R$\&$D Program of China under Grant  No.2020YFC0832505.
	\bibliographystyle{ACM-Reference-Format}
	\balance
	\bibliography{mm2022}
	
%\clearpage
\appendix
%\onecolumn
	\begin{table*}[h]
	%\addtocounter{table}{+7}
	%\centering
	\begin{tabular}{l|c|c|c|c|c|c}
		\toprule
		\multirow{2}*{\textbf{Method}} & \multicolumn{3}{c|}{\textbf{IJB-B 1:1 TAR (\%)}} & \multicolumn{3}{c}{\textbf{IJB-C 1:1 TAR (\%)}}\\
		\cline{2-7} 
		~ & FAR=1e-6 & FAR=1e-5 & FAR=1e-4 &FAR=1e-6 & FAR=1e-5 & FAR=1e-4  \\
		\hline
		Vanilla (Entangled) &\st{39.25} & 88.18 & 93.79 &84.99 & 92.42 & 95.23  \\
		+ Attention  & \st{41.28} & \textbf{90.10} & \textbf{94.21}  &\textbf{87.94} & \textbf{93.61} & \textbf{95.72}\\
		\hline
		Image-based disentangled & \st{41.50}& 88.75 & 93.90& 85.79 & 92.69 & 95.34   \\
		+ Attention (ID) & \st{41.93} & 90.27 &\textbf{ 94.24}&87.37 & 93.57 & 95.74\\
		+ Attention (VA) &\st{43.89} & \textbf{90.36} & \textbf{94.24}& \textbf{88.31} &\textbf{ 93.63} & \textbf{95.75}\\
		\hline
		%\hdashline
		\textbf{Set-based disentangled}& \st{42.05}&88.90&93.82&86.32 & 92.81 & 95.31\\
		+ Attention (ID) & \st{41.03} & 90.19 & 94.16&88.31 & 93.62 & 95.63\\
		\textbf{+ Attention (VA)} & \st{43.61} & \textbf{90.41} & \textbf{94.18}&\textbf{88.40} & \textbf{93.70}& \textbf{95.72}\\
		\bottomrule
	\end{tabular}
	% \vspace{-0.1in}
	\caption{Ablation of disentanglement and quality attention on IJB-C 1:1 verification protocols with ArcFace backbone. The best results are highlighted in bold.}
	\label{tab:disentangle_arc}
\end{table*}

\begin{table*}[ht]
	\centering
	\begin{tabular}{l|c|c|c|c|c|c}
		\toprule
		\multirow{2}*{\textbf{Method}} & \multicolumn{3}{c|}{\textbf{IJB-B 1:1 TAR (\%)}} & \multicolumn{3}{c}{\textbf{IJB-C 1:1 TAR (\%)}}\\
		\cline{2-7} 
		~ & FAR=1e-6 & FAR=1e-5 & FAR=1e-4 &FAR=1e-6 & FAR=1e-5 & FAR=1e-4  \\
		\hline
		Vanilla (Entangled) &\st{39.25} & 88.18 & 93.79 &84.99 & 92.42 & 95.23\\
		+ VBS  & \st{41.22} & 88.35 & 93.85&86.31 & 92.86 & 95.50\\
		+ VBS + Attention & \st{42.88} & \textbf{89.98} & \textbf{94.28}&\textbf{88.99} & \textbf{93.63} & \textbf{95.84}\\ 
		\hline
		\textbf{Set-based disentangled} &\st{42.05}&88.90&93.82&86.32 & 92.81 & 95.31 \\
		+ VBS (ID) & \st{40.73}& 87.75 & 93.63& 85.22 & 92.27 & 95.19\\ 
		\textbf{+ VBS (VA)} &  \st{46.39} & 90.45 & 94.34&89.06 & \textbf{93.90} & 95.84\\ 
		\textbf{+ VBS (VA) + VQA} & \st{43.49} & \textbf{90.72} & \textbf{94.38}&\textbf{89.31} & 93.82 & \textbf{95.85} \\
		\bottomrule
	\end{tabular}
	% \vspace{-0.1in}
	\caption{Ablation of vocabulary-based burst suppression on entangled and disentangled models with ArcFace backbone. }
	\label{tab:disentangle_burst_arc}
\end{table*}

\section{Ablation on ArcFace backbone}
The VGGFace2 model is adopted for the ablations in Table~1 and Table~2 of the main paper. The observations on the ArcFace-based model are similar and corresponding result comparisons are presented in Table~\ref{tab:disentangle_arc} and Table~\ref{tab:disentangle_burst_arc}.
\subsection{The Impact of Disentanglement and VQA} 
To demonstrate the effectiveness of the proposed light-weighted set-based disentanglement (SD) framework, we compare three kinds of features: the vanilla entangled features, image-based disentangled identity features and the proposed SD identity features. Then we apply quality attention to these features and give the result comparisons in Table~\ref{tab:disentangle_arc}. 
The results of the 1:1 TAR when FAR=1e-6 are not stable on IJB-B dataset as claimed in the main paper. So we apply strikethrough on these results and recommend the reader to focus on the other metrics.

The first observation that can be drawn is that disentanglement improves the feature robustness, and set-based disentangled features usually outperform entangled features and image-based features. 
We can also observe the attention block consistently improves the recognition performance for three features. The attention blocks based on variance features are usually more prominent than identity-based and the entangled-based blocks, which demonstrates the effect of variance-based quality assessment (VQA).

\subsection{The Impact of Variance-Based Burst Suppression.} 

To demonstrate the effectiveness of VBS, we apply it to the vanilla entangled model and SD model.  As can be seen in Table~\ref{tab:disentangle_burst_arc}, the proposed VBS is effective on both the entangled and disentangled models with considerable performance improvement. 
For the SD model, the VBS on identity features gets worse results because the face burstiness is identity-unrelated. 
In addition, VBS is complementary to the attention mechanisms VQA on SD model, and their combination further improves the performance.

\section{1:N identification performance}
The 1:N identification performance on IJB-B and IJB-C is shown in Table~\ref{tab:SOTA_1n}. It can be observed that the proposed set-based disentanglement (SD) and vocabulary-based aggregation (VBA) methods improve the backbone model by large margins. The vocabulary-based burst suppression (VBS) with feature norm as quality scores on the SD framework gets best results with ArcFace backbone.
%Comparatively, the improvements on ArcFace backbone are slighter, we suppose the .
\begin{table*}[ht]
	\centering
	\begin{tabular}{l|c|c|c|c|c|c|c|c}
		\toprule
		\multirow{2}*{\textbf{Method}} & \multicolumn{4}{c|}{\textbf{IJB-B 1:N TPIR}} & \multicolumn{4}{c}{\textbf{IJB-C 1:N TPIR}} \\
		\cline{2-9} 
		~ &FPIR=0.01& FPIR=0.1 & R@1 & R@5& FPIR=0.01& FPIR=0.1 & R@1 & R@5 \\
		\hline
		VGGFace2\cite{cao2018vggface2}&61.07&76.78&88.46&93.46&70.18&79.30&89.67&93.99\\
		\textbf{+ SD + VBA}&\textbf{72.12}&\textbf{84.81}&\textbf{90.46}&\textbf{94.31}& \textbf{78.18}&\textbf{85.76}&\textbf{91.49}&\textbf{94.93}\\
		\hline
		ArcFace \cite{deng2019arcface} &83.31& 93.38&94.50&96.58&90.33&94.52&95.72&97.10\\   %83.31& 93.38&94.50&96.58	 
		\textbf{SD + VBA} &83.60&93.67&94.65&96.64&92.21&94.82&96.03&97.27 \\
		%+ SD + VBS + Norm &\textbf{83.17} &93.46 &94.63 & 96.64&91.96&9460 &95.93&97.32 \\
		\textbf{SD + VBS + Norm} & \textbf{84.01}& \textbf{93.77}& \textbf{94.77}&\textbf{96.66} &\textbf{92.67}&\textbf{95.03} &\textbf{96.16} &\textbf{97.36}\\
		\bottomrule
	\end{tabular}
	% \vspace{-0.1in}
	\caption{Ablation of set-based disentanglement and vocabulary-based aggregation methods on IJB-B and IJB-C 1:N identification protocols. }
	\label{tab:SOTA_1n}
\end{table*}
\end{document}